\documentclass[conference]{IEEEtran}
\IEEEoverridecommandlockouts
\usepackage{cite}
\usepackage{amsmath,amssymb,amsfonts}
\usepackage{algorithm}
\usepackage{graphicx}
\usepackage{textcomp}
\usepackage{xcolor}
\usepackage{stfloats}
\usepackage{booktabs}
\usepackage{multirow}
\usepackage{algorithm}
\usepackage{algpseudocode}
\def\BibTeX{{\rm B\kern-.05em{\sc i\kern-.025em b}\kern-.08em
    T\kern-.1667em\lower.7ex\hbox{E}\kern-.125emX}}
\begin{document}

\title{Center-Oriented Prototype Contrastive Clustering
\thanks{$^\star$Corresponding author. This work was funded in part by the National Natural Science Foundation of China (92470202, 62272468 and 62376252); Key Project of Natural Science Foundation of Zhejiang Province (LZ22F030003); Zhejiang Province Leading Geese Plan (2025C02025, 2025C01056); Zhejiang Province Province-Land Synergy Program (2025SDXT004-3).}
}

\author{\IEEEauthorblockN{1\textsuperscript{st} Shihao Dong}
\IEEEauthorblockA{\textit{School of Computer Science} \\
\textit{Nanjing University of Information Science and Technology}\\
Nanjing, China \\
dongshihao@nuist.edu.cn}
\and
\IEEEauthorblockN{2\textsuperscript{nd} Xiaotong Zhou}
\IEEEauthorblockA{\textit{School of Computer Science} \\
\textit{Nanjing University of Information Science and Technology}\\
Nanjing, China \\
xiaotong\_zhou@nuist.edu.cn}
\and
\IEEEauthorblockN{3\textsuperscript{rd} Yuhui Zheng}
\IEEEauthorblockA{\textit{The State Key Laboratory of Tibetan Intelligence} \\
\textit{Qinghai Normal University}\\
Xining, China \\
zhengyh@vip.126.com}
\and
\IEEEauthorblockN{4\textsuperscript{th} Huiying Xu$^\star$}
\IEEEauthorblockA{\textit{Computer Science and Technology} \\
\textit{Zhejiang Normal University}\\
Jinhua, China \\
xhy@zjnu.edu.cn}
\and
\IEEEauthorblockN{5\textsuperscript{th} Xinzhong Zhu}
\IEEEauthorblockA{\textit{Computer Science and Technology} \\
\textit{Zhejiang Normal University}\\
Jinhua, China \\
zxz@zjnu.edu.cn}
}

\maketitle

\begin{abstract}
Contrastive learning is widely used in clustering tasks due to its discriminative representation. However, the conflict problem between classes is difficult to solve effectively. Existing methods try to solve this problem through prototype contrast, but there is a deviation between the calculation of hard prototypes and the true cluster center. To address this problem, we propose a center-oriented prototype contrastive clustering framework, which consists of a soft prototype contrastive module and a dual consistency learning module. In short, the soft prototype contrastive module uses the probability that the sample belongs to the cluster center as a weight to calculate the prototype of each category, while avoiding inter-class conflicts and reducing prototype drift. The dual consistency learning module aligns different transformations of the same sample and the neighborhoods of different samples respectively, ensuring that the features have transformation-invariant semantic information and compact intra-cluster distribution, while providing reliable guarantees for the calculation of prototypes. Extensive experiments on five datasets show that the proposed method is effective compared to the SOTA. Our code is published on https://github.com/LouisDong95/CPCC.
\end{abstract}

\begin{IEEEkeywords}
self-supervised learning, contrastive learning, representation learning, deep clustering.
\end{IEEEkeywords}

\section{Introduction}
\label{sec:intro}
In the data-driven world, deep clustering, as a basic tool for exploring the intrinsic structure and patterns of high-dimensional data, has been a research hotspot in the fields of deep learning and data mining. In recent years, contrastive learning~\cite{he2020momentum, chen2020simple} has been widely used in clustering tasks due to its ability to efficiently learn feature representations of data without expensive labels by using the information of the data itself as training signals.

\begin{figure}
    \centering
    \includegraphics[width=0.7\linewidth]{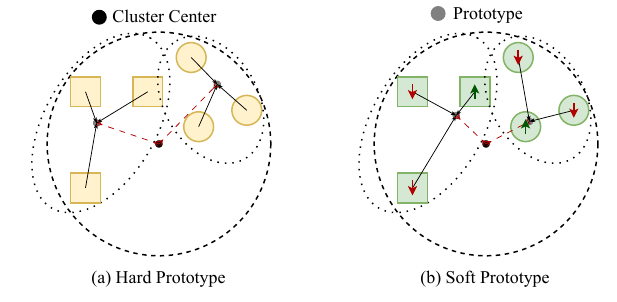}
    \caption{\textbf{Our motivation}. The circular squares represent samples with different transformations. The prototypes are calculated by random sampling. The hard prototype calculation treats each sample equally, causing the calculated prototype to deviate from the true cluster center. The soft prototype assigns higher weights to samples close to the cluster center and reduces the weights of distant points, so that the calculated prototype is closer to the true cluster center}
    \label{fig_idea}
\end{figure}

Although some works~\cite{van2020scan, li2021contrastive, niu2022spice, dang2021nearest} have achieved good performance by combining deep clustering with contrastive learning, contrastive learning treats the same samples and their transformations as positive pairs, while the remaining samples are treated as negative pairs. Negative pairs contain samples of the same class with the same semantic information, which inevitably leads to inter-class conflicts, thereby destroying the discriminative information. Therefore, how to alleviate false pairs has always been a hot topic of concern in contrastive learning. Most existing methods alleviate the impact of false pairs by reducing the weight of false negative pairs and mining false positive pairs. GCC~\cite{zhong2021graph} constructs positive pairs through the adjacency graph, and NNM~\cite{dang2021nearest} constructs positive pairs by combining local and global neighbors. GDCL~\cite{zhao2021graph} deleted false negative samples that are similar to the original samples through a bias elimination strategy. TCL~\cite{li2022twin} removed the false negative samples from the contrastive loss denominator. HSAN~\cite{liu2023hard} reduced the weight between easy sample pairs and increased the weight between difficult sample pairs. DIVIDE~\cite{Lu0YPH024} reduces the weight of false positive samples through random walks. Although conflicts between classes are effectively mitigated, it does not fundamentally solve this problem. To avoid inter-class conflicts, PCL~\cite{0001ZXH21} calculates $K$ prototypes and considers the sample and its corresponding prototype as a positive pair, while the remaining prototypes are considered negative samples. ProPos~\cite{HuangCZS23} calculates $2K$ prototypes in pairs through data augmentation, and considers prototypes of the same class as positive pairs, which not only avoids class conflicts but also greatly improves the efficiency of contrastive learning. Although the prototype contrast method effectively avoids inter-class conflicts, it treats samples equally when calculating the prototype, which inevitably introduces noise, causing the prototype estimation to deviate from the true cluster center and inaccurate prototype calculation, as shown in the Fig.~\ref{fig_idea}.

To address the above problem, we propose a $\textbf{C}$enter-oriented $\textbf{P}$rototype $\textbf{C}$ontrastive $\textbf{C}$lustering method, termed CPCC. Our method consists of soft prototype contrastive (SPC) and dual consistency learning (DCL). Specifically, SPC considers prototypes from the same category as positive pairs and prototypes from different categories as negative pairs. The prototype is calculated by taking the square of the probability that each sample belongs to its cluster center as the weight, so that high-confidence samples are assigned higher weights, otherwise, they are assigned lower weights. This strategy can not only avoid category conflicts but also reduce the deviation of prototypes from cluster centers. In addition, DCL learns the consistency of the features of the same sample under different transformations and the consistency of the features between different samples and their neighbors, resulting in further compression of the intra-class feature space. The contributions are summarized as follows:
\begin{itemize}
\item We propose a novel prototype contrastive clustering method to avoid inter-cluster conflicts and improve intra-cluster compactness through prototype contrast and consistency learning.
\item Compared with hard prototypes, we propose a soft prototype construction method that can avoid prototype drift.
\item Extensive experimental results on five benchmark datasets demonstrate the effectiveness of the proposed method compared with the existing SOTA.
\end{itemize}

\section{Related Work}
\subsection{Deep Clustering}
Deep clustering utilizes deep learning techniques to perform clustering tasks. It guides the learning process through the features and structure of the data itself without relying on human intervention. According to the network model, deep clustering can be divided into generative network-based and discriminative network-based deep clustering. 1) Generative network-based, such as DEC~\cite{xie2016unsupervised} realized clustering on embedded features by reconstructing the data through AE. VaDE~\cite{JiangZTTZ17} sampled from Gaussian Mixture Models distributions which not only realized clustering but also generated new samples. ClusterGAN~\cite{mukherjee2019clustergan} realized clustering and generated samples through the generative adversarial network. 2) Discriminative networks-based, such as SCAN~\cite{van2020scan} pre-trained a pair of siamese networks with shared weights by contrastive learning and combined them with neighbor assignment consistency to achieve clustering. NNM~\cite{dang2021nearest} extended neighbor assignment consistency to local neighbors based on SCAN. CC~\cite{li2021contrastive} improved the discrimination of the cluster space through bi-level contrasts. SPICE~\cite{niu2022spice} combined contrastive learning and pseudo-labeling for semi-supervised training of networks.

\subsection{Contrastive Learning}
Recently, contrastive learning has received much attention for its ability to learn discriminative feature representations from unlabeled data. Such methods are usually combined with a specific task to implement clustering based on MoCo~\cite{he2020momentum} or SimCLR~\cite{chen2020simple}, whose loss functions are usually implemented by InfoNEC or NT-Xnet. However, these two methods usually require a large number of negative samples, which leads to class conflicts. One way to avoid conflicts is to not require negative sample pairs: BYOL~\cite{grill2020bootstrap} is unique in that it does not require negative pairs for contrastive learning, but rather learns the feature representations by maximizing the consistency of the outputs of the online network and the target network, thus avoiding the selection of negative samples. Another way is through prototype contrastive: PCL~\cite{0001ZXH21} is a contrastive framework between samples and their prototypes, and ProPos~\cite{HuangCZS23} is a contrastive sample between prototypes. It takes prototypes of the same type as positive pairs and prototypes of different types as negative pairs, which not only avoids inter-class conflicts but also reduces computational complexity.

\begin{figure}[tp]
    \centering
    \includegraphics[width=1\linewidth]{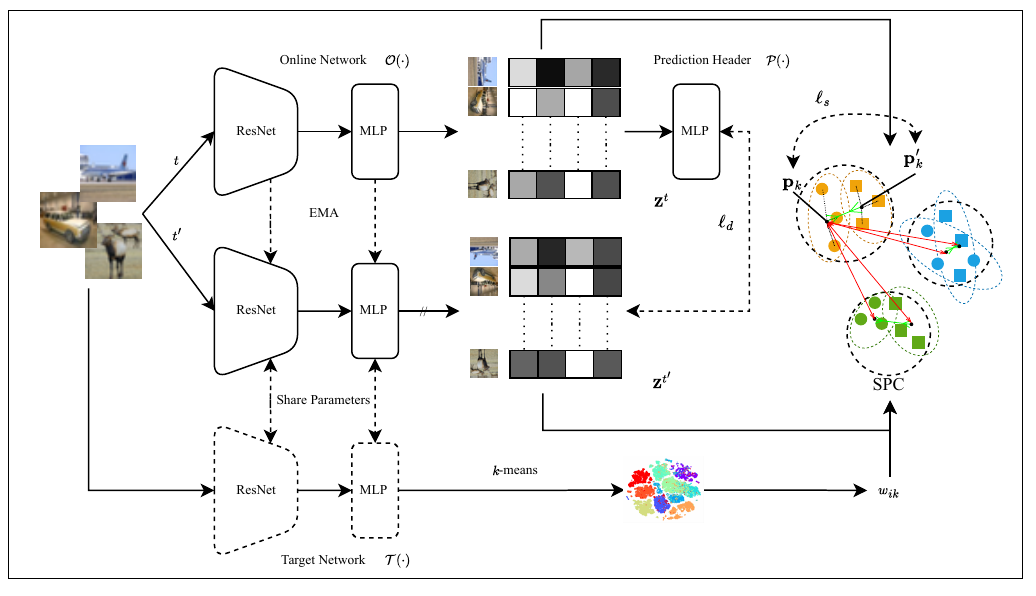}
    \caption{\textbf{Illustration of CPCC framework.} Our framework consists of an online network $\mathcal{O}(\cdot)$ and a target network $\mathcal{T}(\cdot)$. The target network synchronizes parameters with the online network through a moving average strategy. First, the cluster center of the original data set is obtained through the $\mathcal{T}(\cdot)$, and the weight $\mathbf{w}_i$ of each sample is calculated based on the distance between the feature and its center. Secondly, the features $\mathbf{Z}^t$ and $\mathbf{Z}^{t'}$ of different transformed samples are obtained through the $\mathcal{O}(\cdot)$ and the $\mathcal{T}(\cdot)$ respectively, $2K$ prototypes of different transformations are calculated based on the features and its weights, and the network is trained through SPC loss and DCL loss}
    \label{pipeline_fig}
\end{figure}

\section{Method}
\label{sec:Method}
\subsection{Soft Prototype Contrastive}
Given a dataset $\mathcal{X}=\{\mathbf{x}_i\}_{i=1}^N \in\mathbb{R}^{N\times D}$ and a set of transformations $T$, contrastive learning constructs pairs of samples by data augmentation $\{\mathbf{X}^t, \mathbf{X}^{t'}\}, t, t' \in T$. Where different transformations of the same sample are considered positive pairs and the remaining samples are considered negative pairs. The discrimination of each sample representation can be effectively increased by increasing the similarity of positive pairs and decreasing the similarity of negative pairs. The formula is as follows:
\begin{equation}\label{eq_contrastive}
    \ell_{i} = -\frac{1}{N}\sum_{i=1}^N\log\frac{e^{(\mathbf{z}_i^t)^\top \mathbf{z}_i^{t'}/\tau}}
    {\sum_{j=1,j\neq i}^{N} [e^{(\mathbf{z}_i^t)^\top \mathbf{z}_j^t/\tau} + e^{(\mathbf{z}_i^t)^\top \mathbf{z}_j^{t'}/\tau}]}.
\end{equation}

Although contrastive learning methods achieve high-quality feature representation capability by this property, the false negative pairs in the denominator of Eq.~\eqref{eq_contrastive} will lead to inter-class conflicts and affect the final result of clustering. To this end, we propose an SPC module to avoid this conflict. First, we obtain the feature $\mathbf{Z}$ of the sample through the target network $\mathcal{T}(\cdot)$, and perform $k-$means on $\mathbf{Z}$ to initialize the cluster center $\{\boldsymbol{\mu}_i\}_{i=1}^K$ of each class. Then, the soft assignment $\mathbf{Q}$ between each sample $\mathbf{Z}$ and the cluster centers $\boldsymbol{\mu}$ can be calculated from the student's $t$-distribution:
\begin{equation}\label{eq_soft}
    q_{ik} = \frac{(1+\parallel \mathbf{z}_i-\boldsymbol{\mu}_k\parallel^2/\alpha)^{-\frac{\alpha+1}{2}}}
    {\sum_{k'}(1+\parallel \mathbf{z}_i-\boldsymbol{\mu}_{k'}\parallel^2/\alpha)^{-\frac{\alpha+1}{2}}},
\end{equation}
where $q_{ik}$ is denoted as the probability that the $i$-th sample belongs to the $k$-th cluster, $\alpha = 1$ for all experiments. We computed the weights by raising $\mathbf{Q}$ to quadratic such that features close to the cluster center are assigned larger weights and features far from the cluster center are assigned smaller weights:
\begin{equation}\label{eq_weight}
    w_{ik} = \frac{q^2_{ik}/f_k}{\sum_{k'}q^2_{ik'}/f_{k'}},
\end{equation}
where $f_k = \sum_i q_{ik}$ are soft cluster frequencies. 
We estimate the prototypes of each class through mini-batch samples $\mathcal{B}$. Due to the presence of $\mathbf{W}$, the drift between the prototype and the cluster center caused by random sampling is avoided. The prototypes $\mathbf{P}$ and $\mathbf{P}'$ can be obtained by the following formula:
\begin{equation}\label{eq_prototypes}
    \mathbf{p}_k = \frac{\sum_{i=1}^{|\mathcal{B}|}w_{ik}\mathcal{O}(\mathbf{x}_i^t)}{\parallel\sum_{i=1}^{|\mathcal{B}|}w_{ik}\mathcal{O}(\mathbf{x}_i^t)\parallel_2},
    \mathbf{p}'_k = \frac{\sum_{i=1}^{|\mathcal{B}|}w_{ik}\mathcal{T}(\mathbf{x}_i^{t'})}{\parallel\sum_{i=1}^{|\mathcal{B}|}w_{ik}\mathcal{T}(\mathbf{x}_i^{t'})\parallel_2}.
\end{equation}

After we get the $2K$ prototypes $\{\mathbf{p}_1, \mathbf{p}_2, ..., \mathbf{p}_K\}$ and $\{\mathbf{p}'_1, \mathbf{p}'_2, ..., \mathbf{p}'_K\}$, the soft prototype contrastive loss is calculated as follows:
\begin{equation}\label{eq_spc}
    \ell_{s} = -\frac{1}{K}\sum_{k=1}^K\log\frac{e^{(\mathbf{p}_k)^\top\mathbf{p}_k'/\tau}}
    {\sum_{j=1,j\neq k}^K [e^{(\mathbf{p}_k)^\top\mathbf{p}_j/\tau}+e^{(\mathbf{p}_k)^\top\mathbf{p}_j'/\tau}]},
\end{equation}
where the temperature parameter $\tau$ is used to adjust the scale of similarity. With $\ell_{s}$, similar prototypes of the same class are close to each other in space, while prototypes of different classes are far away from each other. This avoids inter-class conflicts and prototype drift, and improves the distinction between categories and stability during training.

\subsection{Dual Consistency Learning}
The same sample after different transformations should be consistent in the feature space, and the consistency of the features allows the network to ignore some details and learn the semantic information that is invariant to the transformations. In addition, the neighbors of different samples should also be close in the feature space. Therefore, we not only consider different transformations of the same sample as positive pairs, but also consider the corresponding neighbors as positive pairs. The dual consistency learning loss is defined as follows:
\begin{equation}\label{eq_dcl}
\begin{split}
    \ell_{d} &= \frac{1}{2N}\sum_i^N(\parallel \mathcal{P}(\mathbf{z}_i^t) - \mathbf{z}_i^{t'}\parallel^2_2 + \parallel\mathcal{P}(\mathbf{z}_i^{t'} + \sigma\epsilon) - \mathbf{z}_i^t\parallel^2_2)\\
    &=2-\frac{1}{N}\sum_i^N({<\mathcal{P}(\mathbf{z}_i^t),\mathbf{z}_i^{t'}>} +
    {<\mathcal{P}(\mathbf{z}_i^{t'} + \sigma\epsilon) , \mathbf{z}_i^t>}),
\end{split}
\end{equation}
where $\epsilon\sim\mathcal{N}(0,\mathbf{I})$, $\sigma$ is used to control the range of neighbors. $\mathbf{z}_i + \sigma\epsilon$ denotes the neighborhood sampling for $\mathbf{z}_i$. The first part of Eq.~\eqref{eq_dcl} represents the consistency between the sample prediction and its transformation, and the second part represents the consistency between the sample's neighborhood prediction and its transformation. The two constrain transformation invariance and intra-class compactness respectively. Consistency learning not only aligns features of different transformations, but also provides high-confidence pseudo-labels for prototypes.

\subsection{Model Training}
In summary, the total loss of CPCC is defined as:
\begin{equation}\label{eq_loss}
    \ell = \ell_{d} + \lambda\ell_{s},
\end{equation}
where $\lambda$ is a trade-off between $\ell_{s}$ and $\ell_{d}$. The training is divided into two stages. Since the clustering results are unreliable in the early stage of training, only $\ell_d$ is used in the pre-training stage, and the contrastive stage is trained by Eq.~(\ref{eq_loss}). The parameters $\boldsymbol{\theta}_{\mathcal{O}}$ of $\mathcal{O}(\cdot)$ and the parameters
$\boldsymbol{\theta}_{\mathcal{T}}$ of $\mathcal{T}(\cdot)$ is updated by EMA as follows:
\begin{equation}\label{eq_ma}
    \boldsymbol{\theta}_{\mathcal{T}} = m\boldsymbol{\theta}_{\mathcal{T}} + (1-m)\boldsymbol{\theta}_{\mathcal{O}},
\end{equation}
Where $m\in[0, 1)$ is the coefficient that controls the rate of updating. The procedure for the optimization algorithm is summarized in Algorithm \ref{alg}.

\begin{algorithm}[!tb]
    \caption{CPCC}
    \label{alg}
    \begin{algorithmic}[1]
        \Statex \textbf{Input:} Dataset $\mathcal{X}$; data transformation $t, t'\sim\mathcal{T}$; epochs $E$, batch size $B$; networks $\mathcal{O}(\cdot), \mathcal{T}(\cdot)$.
        \Statex \textbf{Output:} Clustering result.
        \For{epoch = 1 to $E$}
            \State Compute the features $\mathbf{Z} = \mathcal{T}(\mathbf{X})$.
            \State Perform $k$-means on $\mathbf{Z}$ and initialize cluster center $\boldsymbol{\mu}$.
            \State Calculate the weights $\mathbf{W}$ by Eq.~\eqref{eq_soft} and Eq.~\eqref{eq_weight}.
            \For{batch = 1 to $\lfloor \frac{|\mathbf{X}|}{B}\rfloor$}
               \State Transformation of each sample $\mathbf{X}^{t}, \mathbf{X}^{t'}$.
               \State Compute the features $\mathbf{Z}^{t} = \mathcal{O}(\mathbf{X}^{t}), \mathbf{Z}^{t'} = \mathcal{T}(\mathbf{X}^{t'})$.
               \State Compute the prototypes $\mathbf{P}, \mathbf{P}'$ by Eq.~\eqref{eq_prototypes}.
               \State Calculate loss $\ell_{s}, \ell_{d}$ by  Eq.~\eqref{eq_spc} and Eq.~\eqref{eq_dcl}.
               \State Update parameters $\boldsymbol{\theta}_{\mathcal{O}}, \boldsymbol{\theta}_{\mathcal{T}}$ by Eq.~\eqref{eq_loss} and Eq.~\eqref{eq_ma}.
            \EndFor
        \EndFor
        \State Perform $k$-means on $\mathbf{Z}$ to obtain the final clustering result.
    \end{algorithmic}
\end{algorithm}

\section{EXPERIMENTS}
\subsection{Datasets and Evaluation Metrics}
We conducted experiments on five benchmark datasets, including CIFAR-10~\cite{krizhevsky2009learning}, CIFAR-20~\cite{krizhevsky2009learning}, STL-10~\cite{coates2011analysis}, ImageNet-10~\cite{chang2017deep}, and ImageNet-Dogs~\cite{chang2017deep}. 
Datasets summary in Table~\ref{tab_dataset}.
Three widely used clustering metrics including Normalized Mutual Information (NMI), Clustering Accuracy (ACC), and Adjusted Rand Index (ARI) are utilized to evaluate our method. Higher scores indicate better clustering performance.

\begin{table}
    \centering
    \caption{Summary of the Datasets}
    \begin{tabular}{lcccc}
    \toprule
        Dataset & Split & Samples & Image size & Classes\\
    \midrule
        CIFAR-10 & Train + Test & 60,000 & $32\times32$ & 10\\
        CIFAR-20 & Train + Test & 60,000 & $32\times32$ & 20\\
        STL-10 & Train + Test & 13,000 & $96\times96$ & 10\\
        ImageNet-10 & Train & 13,000 & $96\times96$ & 10\\
        ImageNet-Dogs & Train & 19,500 & $96\times96$ & 15\\
    \bottomrule
    \end{tabular}
    \label{tab_dataset}
\end{table}

\subsection{Implementation Details}
In our framework, we used ResNet-34 as the backbone to compare with other methods for fairness. To avoid uncertain cluster assignment in the early stage, pre-training is performed by $\ell_{d}$, and $\ell_{s}$ is added after 50 epochs, the learning rate keeps decaying from $0.05$ to $0$. The parameters $\alpha$ = 1, $\tau = 0.5$, $\lambda =0.1$, $\sigma=0.001$ and $m \in [0,1)$ respectively. The experiment was based on multiple averages as the final result and experimental environment contains one desktop computer with Intel Xeon(R) Silver 4310 CPU, two NVIDIA GeForce RTX 4090 GPUs, 128GB RAM, and coded with the PyTorch deep learning platform on the Ubuntu 22.04 operating system.

\begin{table*}[tb]
    \centering
    \caption{The clustering performance on five Datasets}
    \begin{tabular}{lcccccccccccccccccc}
    \toprule
         \multirow{2}*{Methods} & \multicolumn{3}{c}{CIFAR-10} & \multicolumn{3}{c}{CIFAR-20} & \multicolumn{3}{c}{STL-10}& \multicolumn{3}{c}{ImageNet-10} & \multicolumn{3}{c}{ImageNet-Dogs}\\
         \cmidrule(lr){2-4}\cmidrule(lr){5-7}\cmidrule(lr){8-10}\cmidrule(lr){11-13}\cmidrule(lr){14-16}
         ~ & NMI & ACC & ARI & NMI & ACC & ARI & NMI & ACC & ARI & NMI & ACC & ARI & NMI & ACC & ARI\\
    \midrule
         $k$-means~\cite{macqueen1967some} & 0.087 & 0.229 & 0.049 & 0.084 & 0.130 & 0.028 & 0.125 & 0.192 & 0.061 & 0.119 & 0.241 & 0.057 & 0.055 & 0.105 & 0.020\\
         SC~\cite{zelnik2004self} & 0.103 & 0.247 & 0.085 & 0.090 & 0.136 & 0.022 & 0.098 & 0.159 & 0.048 & 0.151 & 0.274 & 0.076 & 0.038 & 0.111 & 0.013\\
         AC~\cite{gowda1978agglomerative} & 0.105 & 0.228 & 0.065 & 0.098 & 0.138 & 0.034 & 0.239 & 0.332 & 0.140 & 0.138 & 0.242 & 0.067 & 0.037 & 0.139 & 0.021\\
         NMF~\cite{cai2009locality} & 0.081 & 0.190 & 0.034 & 0.079 & 0.118 & 0.026 & 0.096 & 0.180 & 0.046 & 0.132 & 0.230 & 0.065 & 0.044 & 0.118 & 0.016\\
    \midrule
         AE~\cite{bengio2006greedy} & 0.239 & 0.314 & 0.169 & 0.100 & 0.165 & 0.048 & 0.250 & 0.303 & 0.161 & 0.210 & 0.317 & 0.152 & 0.104 & 0.185 & 0.073\\
         DAE~\cite{vincent2010stacked} & 0.251 & 0.297 & 0.163 & 0.111 & 0.151 & 0.046 & 0.224 & 0.302 & 0.152 & 0.206 & 0.304 & 0.138 & 0.104 & 0.190 & 0.078\\
         DCGAN~\cite{RadfordMC15} & 0.265 & 0.315 & 0.176 & 0.120 & 0.151 & 0.045 & 0.210 & 0.298 & 0.139 & 0.225 & 0.346 & 0.157 & 0.121 & 0.174 & 0.078\\
         DeCNN~\cite{ZeilerKTF10} & 0.240 & 0.282 & 0.174 & 0.092 & 0.133 & 0.038 & 0.227 & 0.299 & 0.162 & 0.186 & 0.313 & 0.142 & 0.098 & 0.175 & 0.073\\
         VAE~\cite{KingmaW13} & 0.245 & 0.291 & 0.167 & 0.108 & 0.152 & 0.040 & 0.200 & 0.282 & 0.146 & 0.193 & 0.334 & 0.168 & 0.107 & 0.179 & 0.079\\
         JULE~\cite{yang2016joint} & 0.192 & 0.272 &0.138 &0.103 &0.137 &0.033 &0.182 &0.277 &0.164 &0.175& 0.300& 0.138& 0.054& 0.138 & 0.028\\
         DEC~\cite{xie2016unsupervised} & 0.257& 0.301& 0.161& 0.136& 0.185& 0.050 &0.276& 0.359 &0.186 & 0.282& 0.381& 0.203 &0.122 &0.195 &0.079 \\
         DAC~\cite{chang2017deep} & 0.396 & 0.522 &0.306& 0.185 &0.238 &0.088 &0.366 &0.470& 0.257 & 0.394 &0.527 &0.302 &0.219 &0.275 &0.111\\
         DCCM~\cite{wu2019deep} & 0.496 &0.623 &0.408& 0.285& 0.327 &0.173 &0.376 &0.482 &0.262& 0.608& 0.710 &0.555 &0.321 &0.383 &0.182 \\
         IIC~\cite{ji2019invariant} & 0.513 & 0.617 & 0.411 & 0.225 & 0.257 & 0.117 & 0.431 & 0.499 & 0.295 & - & - & - & - & - & - \\
         PICA~\cite{huang2020deep} & 0.591 & 0.696 &0.512 &0.310& 0.337& 0.171 &0.611& 0.713 &0.531 &0.802 &0.870 &0.761 &0.352& 0.352 &0.201\\
    \midrule
         SCAN~\cite{van2020scan} & 0.797 & 0.883 & 0.772 & 0.486 & 0.507 & 0.333 & 0.698 & 0.809 & 0.646& - & - & - & 0.612 & 0.593 & 0.457\\
         GCC~\cite{zhong2021graph} & 0.764 & 0.856 &0.728 &0.472 &0.472 &0.305 &0.684& 0.788 &0.631&0.842& 0.901 &0.822& 0.490& 0.526 &0.362\\
         NNM~\cite{dang2021nearest} & 0.748 & 0.843 & 0.709 & 0.484 & 0.477 & 0.316 & 0.694 & 0.808 & 0.650& - & - & - & 0.604 & 0.586 & 0.449\\
         CC~\cite{li2021contrastive} & 0.705 & 0.790 & 0.637 & 0.431 & 0.429 & 0.266 & 0.764 & 0.850 & 0.726& 0.859 & 0.893 & 0.822 & 0.445 & 0.429 & 0.274\\
         IDFD~\cite{DBLP:conf/iclr/TaoTN21} & 0.711 & 0.815 & 0.663 & 0.426 & 0.425 & 0.264 & 0.643 & 0.756 & 0.575& 0.898 & 0.954 & 0.901 & 0.546 & 0.591 & 0.413\\
         SPICE~\cite{niu2022spice} & 0.734 & 0.838 & 0.705 & 0.448 & 0.468 & 0.294 & {0.817} & {0.908} & {0.812}& 0.828 & 0.921 & 0.836 & 0.572 & 0.646 & 0.479\\
         TCL~\cite{li2022twin} & 0.819 & 0.887 & 0.780 & 0.529 & 0.531 & 0.357 & 0.799 & 0.868 & 0.757& 0.875 & 0.895 & 0.837 & 0.623 & 0.644 & 0.516\\
         ProPos~\cite{HuangCZS23} & \underline{0.886} & \underline{0.943} & \underline{0.884} & \underline{0.606} & \underline{0.614} & \underline{0.451} & 0.758 & 0.867 & 0.737& \underline{0.896} & \underline{0.956} & \underline{0.906} & \underline{0.692} & \underline{0.745} & \underline{0.627}\\
         RPSC~\cite{00070FY024} & 0.754 & 0.857 & 0.731 & 0.476 & 0.518 & 0.341 & \textbf{0.838} & \textbf{0.920} & \textbf{0.834} & 0.830 & 0.927 & 0.858 & 0.552 & 0.640 & 0.465\\
         Ours & \textbf{0.900} & \textbf{0.950} & \textbf{0.898} & \textbf{0.611} & \textbf{0.618} & \textbf{0.456} & \underline{0.825} & \underline{0.912} & \underline{0.815}& \textbf{0.904} & \textbf{0.962} & \textbf{0.916} & \textbf{0.698} & \textbf{0.749} & \textbf{0.634}\\
    \bottomrule
    \end{tabular}
    \label{tab_comparison}
\end{table*}

\subsection{Performance Comparison}
We report the clustering methods including traditional clustering, $k$-means~\cite{macqueen1967some}, SC~\cite{zelnik2004self}, AC~\cite{gowda1978agglomerative}, NMF~\cite{cai2009locality}; deep clustering methods, AE~\cite{bengio2006greedy}, DAE~\cite{vincent2010stacked}, DCGAN~\cite{RadfordMC15}, DeCNN~\cite{ZeilerKTF10}, VAE~\cite{KingmaW13}, JULE~\cite{yang2016joint}, DEC~\cite{xie2016unsupervised}, DAC~\cite{chang2017deep}, DCCM~\cite{wu2019deep}, IIC~\cite{ji2019invariant}, PICA~\cite{huang2020deep} and contrastive learning based deep clustering methods, SCAN~\cite{van2020scan}, GCC~\cite{zhong2021graph}, NNM~\cite{dang2021nearest}, CC~\cite{li2021contrastive}, IDFD~\cite{DBLP:conf/iclr/TaoTN21}, SPICE~\cite{niu2022spice}, TCL~\cite{li2022twin}, ProPos~\cite{HuangCZS23} and RPSC~\cite{00070FY024}, As shown in Table~\ref{tab_comparison}. the contrastive-based deep clustering methods perform better than general deep clustering methods due to the ability of contrastive learning to learn the discriminative features. In addition, prototype contrastive methods ProPos and CPCC achieve better performance than instance contrastive methods due to the reduced inter-class conflicts. Most importantly, CPCC outperforms the SOTA method in all three metrics on most datasets, further demonstrating the validity of our method.

\begin{table}[!htb]
    \centering
    \caption{Impact of different combinations on performance}
    \begin{tabular}{lcccccc}
    \toprule
         \multirow{2}*{Losses} & \multicolumn{3}{c}{CIFAR-10} &  \multicolumn{3}{c}{CIFAR-20}\\
         \cmidrule(lr){2-4}\cmidrule(lr){5-7}
         ~& NMI & ACC & ARI & NMI & ACC & ARI\\
    \midrule
         BYOL~\cite{grill2020bootstrap} & 0.794 & 0.878 & 0.766 & 0.464 & 0.45 & 0.295\\
         CPCC w/o SPC & 0.813 & 0.881 & 0.765 & 0.567 & 0.522 & 0.380\\
         CPCC w/o DCL & 0.259 & 0.332 & 0.145 & 0.168 & 0.201 & 0.065\\
         -CPCC w/o DCL$_1$ & 0.871 & 0.929 & 0.857 & 0.599 & 0.575 & 0.418\\
         -CPCC w/o DCL$_2$ & 0.813 & 0.881 & 0.765 & 0.565 & 0.522 & 0.380\\
         CPCC w/o W & {0.887} & {0.944} & {0.884} & {0.607} & {0.615} & {0.452}\\
         CPCC  & \textbf{0.900} & \textbf{0.950} & \textbf{0.898} & \textbf{0.611} & \textbf{0.618} & \textbf{0.456}\\
    \bottomrule
    \end{tabular}
    \label{tab_ablation}
\end{table}

\begin{figure}[htb]
    \begin{minipage}[b]{0.32\linewidth}
        \centering
        \centerline{\includegraphics[width=\linewidth]{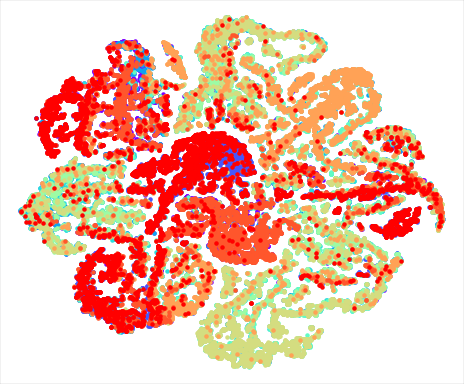}}
        \centerline{(a) CPCC w/o DCL}\medskip
    \end{minipage}
    \begin{minipage}[b]{0.32\linewidth}
        \centering
        \centerline{\includegraphics[width=\linewidth]{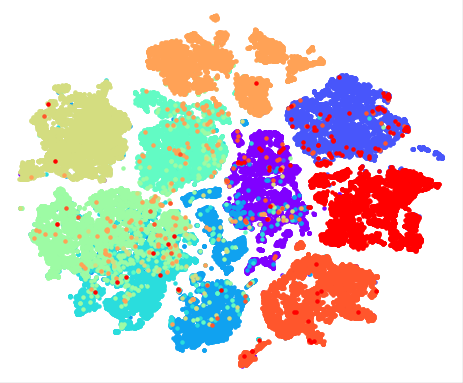}}
        \centerline{(b) CPCC w/o SPC}\medskip
    \end{minipage}
    \begin{minipage}[b]{0.32\linewidth}
        \centering
        \centerline{\includegraphics[width=\linewidth]{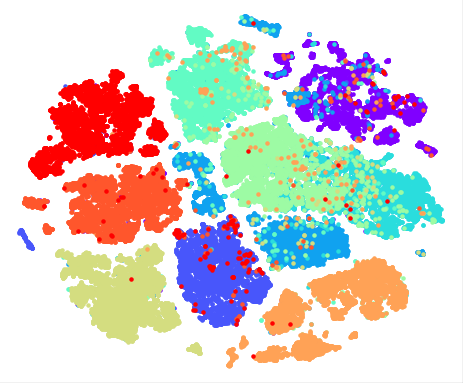}}
        \centerline{(c) CPCC}\medskip
    \end{minipage}
    \caption{$t$-SNE visualization for CPCC on the CIFAR-10 dataset}
    \label{fig_tsne}
\end{figure}
\begin{figure}[htb]
    \begin{minipage}[b]{0.49\linewidth}
        \centering
        \centerline{\includegraphics[width=\linewidth]{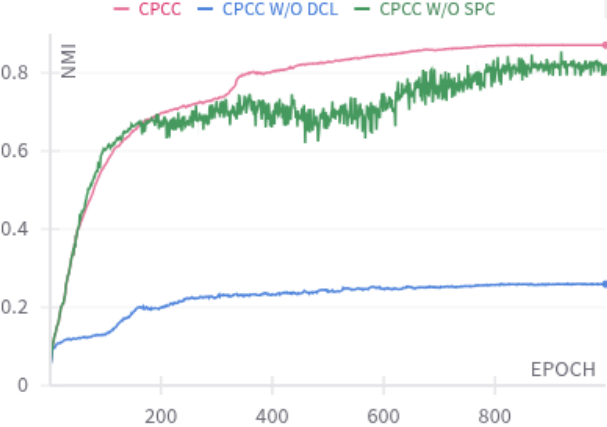}}
        \centerline{(a) NMI}\medskip
    \end{minipage}
    \begin{minipage}[b]{0.49\linewidth}
        \centering
        \centerline{\includegraphics[width=\linewidth]{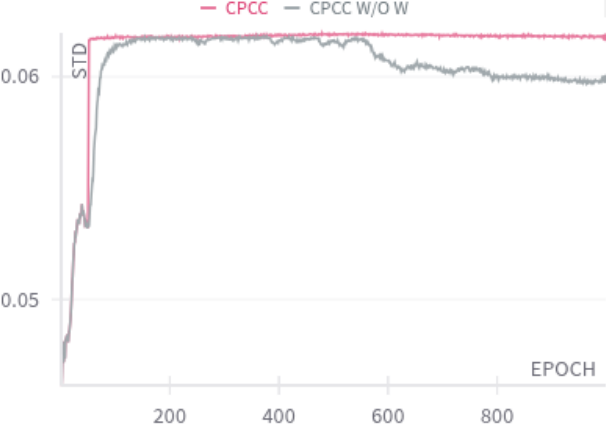}}
        \centerline{(b) STD}\medskip
    \end{minipage}
    \caption{Ablation study on CIFAR-10}
    \label{fig_ablation}
\end{figure}

\subsection{Ablation Study}
\textbf{Effectiveness of DCL}:
To verify the effect of different modules on CPCC, we conducted the experiments shown in Table~\ref{tab_ablation}, and used BYOL~\cite{grill2020bootstrap} as a baseline for comparison, Where DCL$_1$ and DCL$_2$ represent the first and second parts of Eq.~\eqref{eq_dcl} respectively. It is easy to see that both SPC and DCL are indispensable, and without them will lead to performance degradation. In particular, DCL has a great impact on the model's performance, since sample consistency allows the model to learn features that are consistent after sample transformations, which usually contain sample semantic information. We further visualize the features through $t$-SNE, as shown in Fig~\ref{fig_tsne}, which further demonstrates the overall distribution of features. The distribution of CPCC is accurate and compact compared to other features.

\textbf{Effectiveness of SPC}:
In Fig.~\ref{fig_ablation}(a), CPCC w/o SPC achieves good performance but the training process is very unstable, SPC avoids inter-class conflicts, which not only improves the performance but also further improves the stability of the training process. In addition, CPCC w/o W means directly calculating the hard prototype, which has a slightly lower performance than CPCC. We combine the STD (standard deviation) of the features in training to reflect the drift of the prototype. The larger the value of STD, the more uniform the distribution of features in space, and the more accurate the estimation of the prototype. As shown in Fig.~\ref{fig_ablation}(b), the STD of CPCC with weights is larger and more stable, the more uniform the feature distribution, the smaller the prototype drift, therefore the closer to the real clustering center.

\begin{figure}[!h]
    \begin{minipage}[b]{0.49\linewidth}
        \centering
        \centerline{\includegraphics[width=1\linewidth]{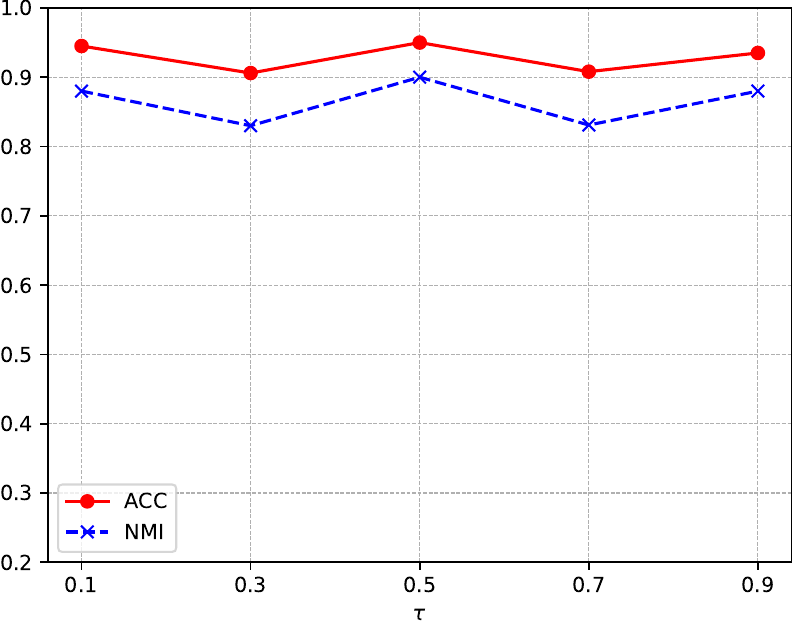}}
        \centerline{(a) Analysis on value of $\tau$}\medskip
    \end{minipage}
    \begin{minipage}[b]{0.49\linewidth}
        \centering
        \centerline{\includegraphics[width=1\linewidth]{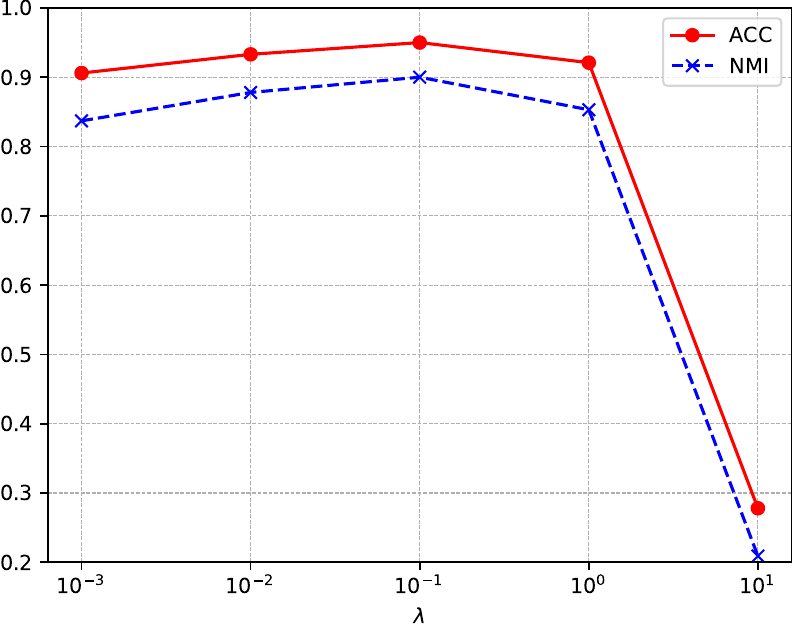}}
        \centerline{(b) Analysis on value of $\lambda$}\medskip
    \end{minipage}
    \caption{Parameter sensitivity analysis on CIFAR-10}
    \label{fig_param}
\end{figure}

\begin{figure}
    \centering
    \includegraphics[width=0.6\linewidth]{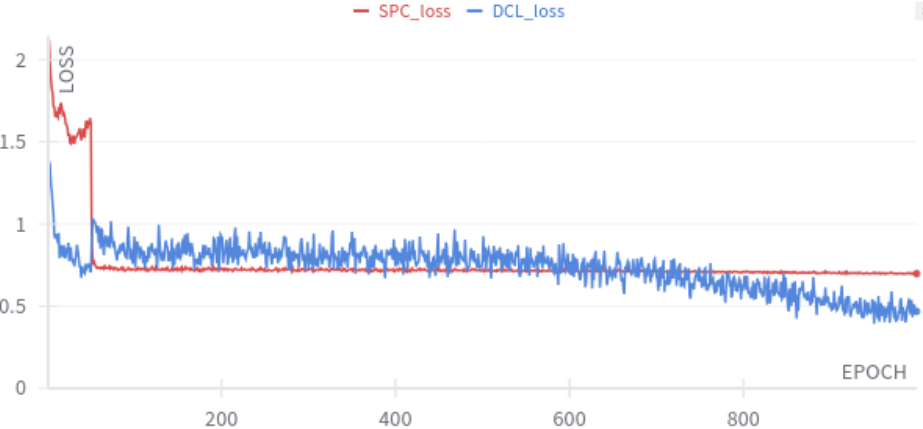}
    \caption{Convergence Analysis on CIFAR-10}
    \label{fig_convergence}
\end{figure}

\subsection{Parameter Sensitivity Analysis}
We further analyze the effect of different values of hyperparameters $\tau$ and $\lambda$ on the results as shown in Fig.\ref{fig_param}. The two correspond to Eq.~\eqref{eq_spc} and Eq.~\eqref{eq_loss}, respectively. The parameter $\tau$ is used to scale the similarity between sample pairs, and its value has little effect on the CPCC method. The parameter $\lambda$ is used to adjust the balance between the $\ell_s$ and the $\ell_d$, and the clustering effect is best when $\lambda=0.1$. For the rest of the hyperparameters, we take empirical values.

\subsection{Convergence Analysis}
Fig.~\ref{fig_convergence} shows the convergence of $\ell_d$ and $\ell_s$. Although $\ell_s$ has been fluctuating, it is generally on a downward trend. After 50 epochs, we added $\ell_s$ to train the network together, $\ell_s$ drops rapidly and remains stable, indicating that the prototypes are well separated and the stability of training is effectively maintained.

\section{Conclusion}
In this paper, we propose a prototype contrast clustering method to solve the inter-class conflict problem of contrastive learning-based clustering methods. By combining soft prototype contrast with feature consistency learning, we can avoid inter-class conflicts and reduce inaccurate prototype calculation caused by prototype drift, effectively improving clustering performance and stability. Experimental results prove the superiority of the proposed method.

\bibliographystyle{IEEEbib}
\bibliography{icme2025references}

\end{document}